# *Grassmannian Discriminant Maps (GDM) for Manifold Dimensionality Reduction with Application to Image Set Classification*

Rui Wang[a], Xiao-Jun Wu[a*], Kai-Xuan Chen[a], Josef Kittler[b]
[a]School of Internet of Things Engineering, Jiangnan University, Wuxi 214122, China
[b]CVSSP, University of Surrey, Guildford, GU2 7XH, UK
{cs_wr, wu_xiaojun}@jiangnan.edu.cn, kaixuan_chen_jnu@163.com, j.kittler@surrey.ac.uk

*Abstract*—**In image set classification, a considerable progress has been made by representing original image sets on Grassmann manifolds. In order to extend the advantages of the Euclidean based dimensionality reduction methods to the Grassmann Manifold, several methods have been suggested recently which jointly perform dimensionality reduction and metric learning on Grassmann manifold to improve performance. Nevertheless, when applied to complex datasets, the learned features do not exhibit enough discriminatory power. To overcome this problem, we propose a new method named Grassmannian Discriminant Maps (GDM) for manifold dimensionality reduction problems. The core of the method is a new discriminant function for metric learning and dimensionality reduction. For comparison and better understanding, we also study a simple variations to GDM. The key difference between them is the discriminant function. We experiment on data sets corresponding to three tasks: face recognition, object categorization, and hand gesture recognition to evaluate the proposed method and its simple extensions. Compared with the state of the art, the results achieved show the effectiveness of the proposed algorithm.**

## I. INTRODUCTION

In the domain of image-set based classification, linear subspaces have been shown to exhibit the ability to provide powerful feature representation [1, 2, 3, 4, 5, 6]. A typical example is a set of face images or a video sequence of face frames which can be approximated by a linear subspace of low dimensionality. The main advantages of using subspace to encode the image set data with large appearance variations [1, 2, 7, 8, 9] are the lower computational complexity and the higher discriminatory power of the learned features. Nevertheless, the distinctive geometric structure of linear subspaces raises the problem of their effective characterization.

From the previous studies [2, 10, 11, 12, 13, 14], it is apparent that the specific geometry spanned by linear subspaces is a class of Riemannian manifold, namely, Grassmann manifold. In consequence, the traditional image processing and classification methods based on Euclidean geometry [15, 16, 17, 18, 19] cannot be applied to the non-Euclidean space of Grassmann manifold directly. In order to overcome this drawback, [10] introduced a metric devised for Grassmann manifolds for the purpose of encoding the non-Euclidean geometry properly.

By applying the well-studied Riemannian metric, some Grassmannian computing methods map the manifold to a high dimensional Hilbert space via a defined kernel function [2, 4, 20, 1]. While the main idea of these methods is to transform the Riemannian manifold to an approximate Euclidean space, the computational cost is always high. Moreover, these methods do not always respect Riemannian manifold properties. Recently, some methods have been proposed that jointly learn a mapping to perform dimensionality reduction and metric learning on Riemannian manifold [1, 29, 30]. The essence of these methods is to directly project the high dimensional Riemannian manifold to a lower dimensional one, which is more discriminative. They have been shown to deliver good performance on some benchmark datasets, as well as offering reduced computational complexity. However, as the linear mapping function being learned on a non-linear Riemannian manifold, it inevitably leads to sub-optimal results.

In the domain of deep learning, the advantages of deep neural networks are manifest in two main aspects: i) powerful feature representation, and ii) effective non-linear training procedure based on backpropagation. In [22], the authors proposed a deep neural network that was built on Grassmann manifold to learn the network parameters. They applied the backpropagation technique based on the Riemannian manifold stochastic gradient descent training procedure. This architecture has greatly surpassed the existing Grassmannian learning methods. However, it computationally very costly.

In this paper, we focus on the problem of Grassmann manifold dimensionality reduction as it plays an important role in computer vision and machine learning. Conventional methods, such as Principal Component Analysis (PCA) and Linear Discriminant Analysis (LDA), are widely used to handle classification tasks in Euclidean spaces. However, the dimensionality of the Riemannian manifold is always high, thus there has been a growing need to extend the Euclidean based dimensionality reduction techniques to Riemannian manifold to extract low dimensional and more discriminative manifold features. The kernel based Riemannian manifold learning methods [2, 4, 20, 21] mainly focus on how to represent the Riemannian manifold, in order to improve the accuracy of classification. As the mapping function of Riemannian manifold to Hilbert space is implicit, the adopted kernel functions are simply used to measure similarity. The associated advantages of Riemannian kernel methods include dimensionality reduction, and the efficiency of approximation computing.

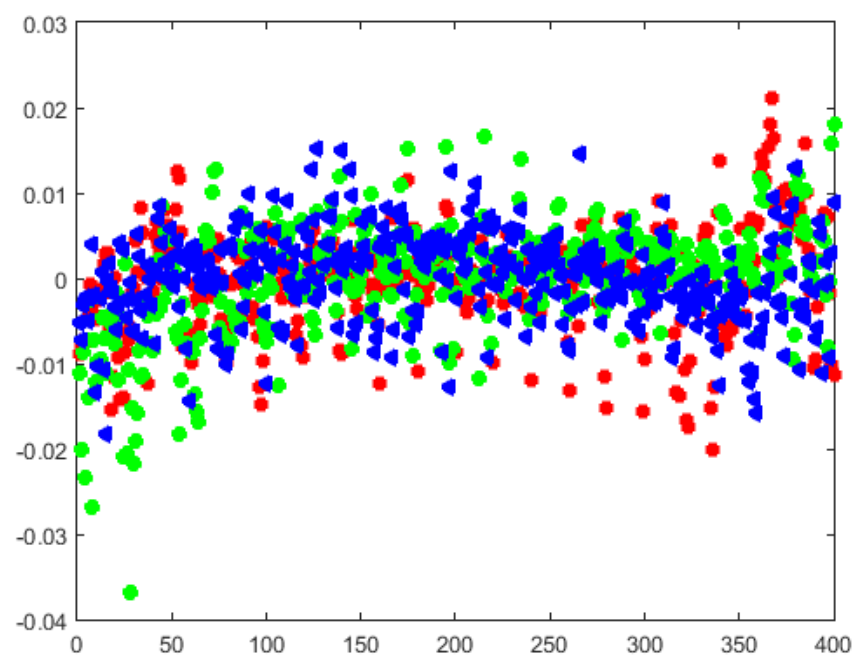

Fig. 1. Original data distribution of three different categories on YTC dataset

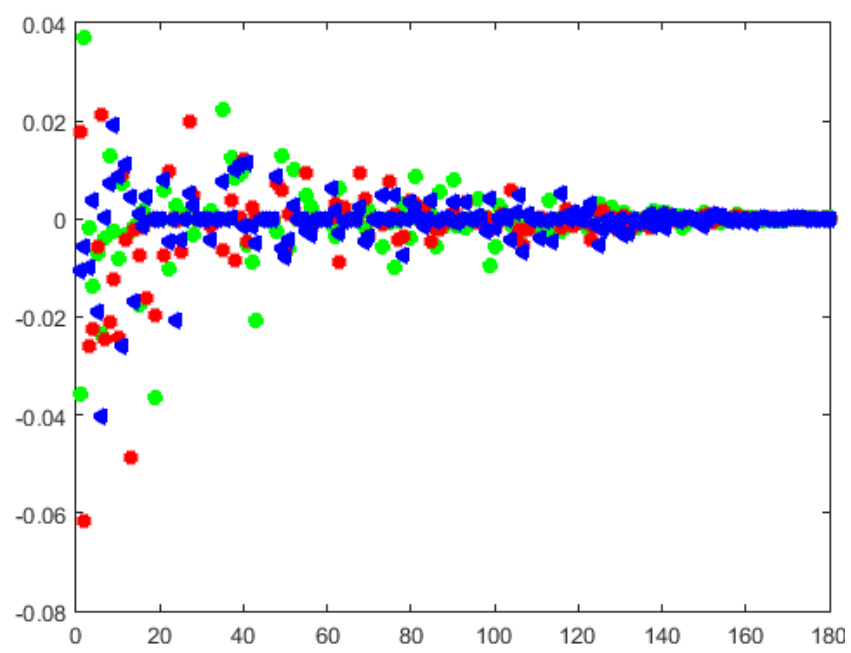

Fig. 2. Data distribution of the generated new subspaces after utilizing PML algorithm

For Riemannian manifold dimensionality reduction problems, it is desirable to make the full use of the geometry of the manifold, so as to extract more discriminative features from complex datasets. Huang et al. in [1] proposed a Grassmann manifold dimensionality reduction method, which jointly performs the process of mapping and metric learning, and generates a low dimensional, more discriminative one. Compared to the kernel based algorithms, by utilizing the manifold geometry, the classification performance of the method surpassed the state-of-the-art methods on some benchmark datasets. But for some complex datasets, like YouTube Celebrities (YTC) [23], it just achieved 66.83% classification accuracy, which demonstrates that the discriminatory information learned by the PML [1] algorithm is not sufficient. Fig.1 shows the data distribution of the original subspaces of the YTC dataset (Three different colors represent three different categories, and in the figure we use class center to represent each class. In addition, the abscissa denotes the dimensionality, and the vertical axis reflects the normalized intensity value) and Fig.2 is the data distribution of the generated new subspaces after applying the PML [1] algorithm. From Fig.1 and Fig.2, we can intuitively see the PML method does not perform very well on the complex dataset in terms of the discriminatory ability of the extracted features.

In this paper, we propose a new discriminant learning method, named Grassmannian Discriminant Maps (GDM). The purpose is to improve the discriminatory ability of the extracted features in the generated new subspaces. The related work is review in Section 2. We overview the PML algorithm and introduce our method in Section 3. Section 4 describes the experiments conducted to validate the method and discusses the results. The conclusions are drawn in Section 5.

## II. Related Work

The traditional methods of Grassmann manifold dimensionality reduction can be divided into three categories: the subspace based manifold learning methods, the kernel based discriminative learning methods, and the methods jointly performing the process of mapping and metric learning.

In the subspace learning methods category, the authors of [24] proposed the Constrained Mutual Subspace Method (CMSM), which looks for a subspace where the canonical correlation of subspace pairs belonging to different classes is small. However, this method shows poor robustness to changes in the target dimensionality of the constrained subspace.

The Discriminant Canonical Correlations (DCC) [25] is also a subspace based learning algorithm, which aims to project the original high dimensional linear subspace to a low-dimensional one to maximize the canonical correlations of inter-class subspace pairs and minimize the canonical correlations of intra-class pairs. However, both CMSM and DCC ignore how the data is distributed in the linear subspaces, which reside on Grassmann manifold, and thus they may produce undesirable results.

In order to overcome the drawbacks of subspace based Grassmannian learning methods, several kernel based discriminative learning algorithms have been proposed. For instance, by designing a projection kernel, which defines a Projection Metric on Grassmann manifold, Grassmann Discriminant Analysis (GDA) [2] embeds the manifold into a high dimensional Hilbert space, and then the Kernel Discriminant Analysis (KDA) [26] is adopted to map it into a lower-dimensional and more discriminative Euclidean space. However, in its mapping process, the local data structure is not considered explicitly, as a result the performance may be compromised. Motivated by this problem, the Grassmannian Graph-embedding Discriminant Analysis (GGDA) [4] algorithm has been proposed to further improve GDA's performance by embedding a more discriminative graph representation of the data. Compared with the subspace based methods, the performance of the kernel based algorithms is better. However, the construction of the kernel matrices is always computational expensive.

The main idea of the above two methods is to transfer the Grassmann manifold into an approximated Euclidean space, so that advanced learning algorithms can be applied for classification. Nevertheless, these kernel based methods fail to exploit the Riemannian manifold geometry, which may lead to sub-optimal results.

Recently, new approaches [1, 27, 28] that directly perform the metric leaning on Riemannian manifold have been suggested. These methods are designed to embed the high-dimensional Riemannian manifold to a lower-dimensional, more discriminative one. Projection Metric Learning (PML) [1] is a popular Grassmann manifold dimensionality-reducing algorithm. This method first defines a mapping on the Grassmann manifold. The discriminant function based on the projection metric is then derived, and finally the Riemannian Conjugate Gradient (RCG) [29] method is applied for optimization. Moreover, the foregoing discussion suggests the PML algorithm has some drawbacks.

In this paper, we introduce a new manifold discriminative learning method, named Grassmannian Discriminant Maps (GDM). The key attribute of GDM is a novel discriminant function based on projection metric, which further improves the discriminatory power of the learned new representation , as a result, the classification performance is improved.

## III. Grassmannian Discriminant Maps

Before we introduce the proposed method, we shall give an overview of the PML algorithm.

### A. *Projection Metric Learning* (*PML*)

Let us consider a $q$ -dimensional linear subspace span $(Y_i)$ on the Grassmann manifold. A general projection mapping: $f: G(q,D) \to G(q,d)$ can be defined as:

$$f\left(Y_i Y_i^T\right) = W^T Y_i Y_i^T W = \left(W^T Y_i\right)\left(W^T Y_i\right)^T \tag{1}$$

where $W \in R^{D\times d}$ is the transformation matrix with column full rank, $Y_i \in R^{D\times q}$ and $Y_i Y_i^T$ represents a point on the original Grassmann manifold. In order to form a valid Grassmann manifold, the QR-decomposition is adopted to maintain the orthogonality of $W^T Y_i$ , and use $W^T Y_i^{'}$ to represent the new linear subspace [1].

For any pair of projection operators $W^T Y_i^{'} Y_i^{'T} W$ and $W^T Y_j^{'} Y_j^{'T} W$ , the projection metric can be defined as:

$$\begin{aligned} & d_P^2\left(W^T Y_i^{'} Y_i^{'T} W, W^T Y_j^{'} Y_j^{'T} W\right) \\ & = 2^{-1/2}\left\|W^T Y_i^{'} Y_i^{'T} W - W^T Y_j^{'} Y_j^{'T} W\right\|_F^2 \\ & = 2^{-1/2} tr\left(P A_{ij} A_{ij}^T P\right) \end{aligned} \tag{2}$$

where $A_{ij} = Y_i^{'} Y_i^{'T} - Y_j^{'} Y_j^{'T}$ and $P = WW^T$ is a rank- $d$ symmetric positive semi-definite (SPSD) matrix.

Based on the projection metric, the discriminant function can be defined as:

$$P^* = \arg\min_P J\left(P\right) = \arg\min_P tr\left(P\left(S_w - \alpha S_b\right)P\right) \tag{3}$$

where $S_w$ reflects the within-class scatter, $S_b$ reflects the between-class scatter, which can be expressed as:

$$S_w = \frac{1}{N_w}\sum_{i=1}^{m}\sum_{j:C_i=C_j} 2^{-1/2}\left(A_{ij}A_{ij}^T\right) \tag{4}$$

$$S_b = \frac{1}{N_b}\sum_{i=1}^{m}\sum_{j:C_i\neq C_j} 2^{-1/2}\left(A_{ij}A_{ij}^T\right) \tag{5}$$

where $N_w$ is the number of pairs of samples from the same class, $N_b$ is the number of pairs of samples from different classes.

***Optimization.*** From the foregoing analysis, we can see the metric learning problem is transformed into optimization for the SPSD matrix $P$ in Eq.3. With its Euclidean gradient $D_P\left(P_k\right) = 2\left(S_w - \alpha S_b\right)P_k$ as input, we can use the nonlinear Riemannian Conjugate Gradient (RCG) algorithm [29] to optimize $P$ . For details of the RCG method, the reader is referred to [1, 28, 29].

### B. *Grassmannian Discriminant Maps* (*GDM*)

In this paper, we propose a new method named Grassmannian Discriminant Maps (GDM) for manifold dimensionality reduction. The core innovation is to design a novel discriminant function, which is defined as:

$$\begin{aligned} P^* &= \arg\min_P J\left(P\right) \\ &= \arg\min_P\left(\exp\left(J_w\left(P\right)\right) + \exp\left(\frac{1}{J_b\left(P\right)}\right) + \lambda\left\|P\right\|_*\right) \end{aligned} \tag{6}$$

where exp is the ordinary matrix exponential operator and $\left\|\cdot\right\|_*$ denotes the matrix nuclear norm. $J_w\left(P\right)$ represents the compactness of within-class scatter, $J_b\left(P\right)$ denotes the between-class scatter. By analogy to Eq.2, they can be defined as:

$$J_w\left(P\right) = tr\left(S_w^*\right) \tag{7}$$

$$J_b\left(P\right) = tr\left(S_b^*\right) \tag{8}$$

where $S_w^*$ and $S_b^*$ are formulated as:

$$S_w^* = P S_w P \tag{9}$$

$$S_b^* = P S_b P \tag{10}$$

and $P$ is the SPSD matrix to be found.

Intuitively, the loss of the first term in Eq. 6 is minimized when $J_w\left(P\right)$ is equal to zero, and that of its second term when $J_b\left(P\right)$ is equal to infinity. The exponential constraint is added to $J_w\left(P\right)$ and $J_b\left(P\right)$ to enhance the non-linear learning ability of the proposed objective function. The purpose of using the nuclear norm is to add the low rank constraint on $P$ to help learn more discriminative feature representation.

***Optimization.*** In Eq.6, we need to optimize the SPSD matrix $P$, however, it is closely related to the linear subspace $Y_i^{'}$. Accordingly, we utilize an iterative solution for the two parameters. First we apply the RCG algorithm to optimize the SPSD matrix $P$ on the manifold spanned by the rank- $d$ SPSD matrices. When $P$ is fixed we use it to optimize $Y_i^{'}$. This procedure can be regarded as an alternating estimation process with the maximum number of iterations set to 15. Recalling the property of the matrix trace operation, Eq.6 can be rewritten as:

$$P^* = \arg\min_P [\exp\left(tr\left(PS_wP\right)\right) + \exp\left(\frac{1}{tr\left(PS_bP\right)}\right) + \lambda \|P\|_*] \quad (11)$$

In the execution of the RCG algorithm, a very important step is to compute the Euclidean gradient of the objective function. But the partial derivative of the third part of Eq.11 does not exist explicitly. Fortunately, it has an approximate mathematical expression:

$$\|P\|_* = \log\det\left(P\right) \quad (12)$$

the partial derivative of which is:

$$\frac{\partial\left(\log\det\left(P\right)\right)}{\partial P} = \left(P^{-1}\right)^T \quad (13)$$

However, $P$ is a SPSD matrix and its inverse form may not exist. In order to get a more accurate result, we add perturbations to $P$ with a regularization term [30, 31]:

$$P \leftarrow P + \frac{v_P}{\alpha} I \quad (14)$$

where $v_p$ represents the sum of all the eigenvalues of $P$, $I$ is an identity matrix, and we let $\alpha = 10^5$ in all the experiments. To initialize $P$, we recall that $P = WW^T$, and initialize the rank-deficient matrix $W$ instead with a truncated identity matrix.

Now, the problem of optimization of the SPSD matrix $P$ has been transformed to one of optimization of $P$ on the Symmetric Positive Definite (SPD) manifold, and the Eq.11 can be rewritten as.

$$P^* = \arg\min_P [\exp\left(tr\left(PS_wP\right)\right) + \exp\left(\frac{1}{tr\left(PS_bP\right)}\right) + \lambda \log\det\left(P\right)] \quad (15)$$

with its Euclidean gradient expressed as:

$$\nabla_P J\left(P\right) = \left(2S_wP\right)\exp\left(tr\left(PS_wP\right)\right) - \exp\left(\frac{1}{tr\left(PS_bP\right)}\right)\left(\frac{\left(2S_bP\right)}{tr\left(PS_bP\right)tr\left(PS_bP\right)}\right) + / \left(P^{-1}\right)^T \quad (16)$$

The main steps of the RCG method are given in Algorithm 1. For the details, please refer to [1, 28, 29].

**Algorithm 1** Riemannian Conjugate Gradient (RCG)

**Input:**

Initialize the SPSD matrix $P$ as $P_0$

$P_0^{'} = P_0 + \frac{v_{P_0}}{\alpha} I$

$H_0 \leftarrow 0$

$P \leftarrow P_0^{'}$

**Repeat:**

1. $H_k \leftarrow -\nabla_P J\left(P_k\right) + \eta\tau\left(H_{k-1}, P_{k-1}, P_k\right)$.
2. Line search along the geodesic with direction $H_k$, to find $P_k = \arg\min_P J(P)$
3. $H_{k-1} \leftarrow H_k$
4. $P_{k-1} \leftarrow P_k$

**Until:** convergence

**Output:** The SPD matrix $P$

*A variant of the proposed GDM*

In order to evaluate the effectiveness of the proposed GDM method, we investigate its variant formulated as follows.

**Variant:**

$$\begin{aligned} P^* &= \arg\min_P J\left(P\right) \\ &= \arg\min_P [\exp\left(J_w\left(P\right)\right) + \log\left(1 + \frac{1}{J_b\left(P\right)}\right) + \lambda \|P\|_*] \end{aligned} \quad (22)$$

which can be rewritten as

$$\begin{aligned} P^* = \arg\min_P [&\exp\left(tr\left(PS_wP\right)\right) \\ &+ \log\left(1 + \frac{1}{tr\left(PS_bP\right)}\right) + / \log\det\left(P\right)] \end{aligned} \quad (23)$$

The Euclidean gradient of Eq.18 feeding into the RCG algorithm is given by:

$$\begin{aligned} \nabla_P J\left(P\right) = &\left(2S_wP\right)\exp\left(tr\left(PS_wP\right)\right) \\ &- \frac{\left(2S_bP\right)}{\left(1 + tr\left(PS_bP\right)\right)} \frac{1}{\left(tr\left(PS_bP\right)\right)} + \lambda\left(P^{-1}\right)^T \end{aligned} \quad (24)$$

## IV. Experiments and Analysis

In order to prove the feasibility and effectiveness of the proposed method, we evaluate it on three tasks: video-based face recognition, set-based object categorization, and hand gesture recognition. For the task of face recognition, we employ the YouTube Celebrities (YTC) [27, 32] dataset. For the object categorization and hand gesture recognition tasks, we use the ETH-80 [27, 31, 32] and Cambridge hand gesture (CHG) [35, 36] benchmark datasets.

### A. *Dataset description and experimental settings*

The challenging YouTube Celebrities data set has 1910 video clips of 47 subjects collected from YouTube. Each clip consists of hundreds of frames, most of which are low resolution and highly compressed. The number of image sets in each subject is also different. In our experiment, we resize each face image to a $20\times 20$ pixel intensity image and in order to eliminate the effects of lighting, the histogram equalization is applied as a pre-processing step.

The ETH-80 data set consists of 8 classes of objects: cows, cups, dogs, horses, pears, tomatoes, cars, and apples with each category containing 10 instances. There are 41 images of different perspectives in each image set. The size of each image is $256\times 256$ . For the sake of consistency with the existing literatures, we resize them to $20\times 20$ and extract their grayscale features.

For the hand gesture recognition task, we utilize the Cambridge hand gesture dataset. This dataset is composed of 9 gesture categories with each class containing 100 image sets. The gestures in this dataset can be divided into 3 hand shapes and 3 motions, and there exists large within-class variations. The size of all the extracted grayscale images are resized to $20\times 20$ .

In all experiments, the $i$ -th image set can be formed as: $X_i=\left[x_1,x_2,\ldots,x_{n_i}\right]$ , where $x_k\in R^{D\times 1}$ is the vectorized representation of the $k$ -th image. After applying the Eigen-Decomposition algorithm to $X_iX_i^T$ , this image set is represented by a linear subspace, which is spanned by $q$ largest eigenvectors. These eigenvectors also compose the orthonormal basis matrix $Y_i$ , which can span a valid Grassmann manifold. The optimal value of $q$ is determined by cross-validation.

We conducted ten-folds cross validation experiments on the three data sets and used ten randomly selected gallery/probe combinations. As for the YTC dataset, each person had three randomly chosen image sets for training and six for testing. For the ETH-80 dataset, we randomly chose five instances in each category for gallery and the other five for probes. When make experiments on CHG dataset, we randomly selected ten image sets in each category for training and another ten for testing.

### B. *Comparative methods and experimental settings*

To evaluate the proposed method, we compare our method with some existing Riemannian manifold learning alternatives, such as Manifold-to-Manifold Distance (MMD) [35], Affine Hull based Image Set Distance (AHISD) [36], SPD Manifold Learning (SPDML) [28] based on Stein divergence [4], Log-Euclidean Metric Learning (LEML) [27], and Projection Metric Learning (PML) [1]. MMD clusters each image set into multiple linear local models and represents each model by a linear subspace. Accordingly, the manifold distance is transformed into subspace distance, which can be computed easily and precisely. AHISD is a subspace-based method, which considers each image set as a convex geometric region. The distance between any two convex regions can be computed by least squares. When it comes to the SPD manifold dimensionality reduction methods, SPDML is a classical algorithm, which jointly performs dimensionality reduction and metric learning directly on the SPD manifold. However, when the dimensionality of the SPD matrix is high, SPDML tends to poor time efficiency. In order to overcome this shortcoming, the LEML algorithm has been proposed which directly performs metric learning and dimensionality reduction on SPD matrices logarithms. To simplify the description, we use GDM-exp to represent the proposed GDM algorithm, and GDM-mix its variant. In our experiments, we also compare the proposed method with its simpler variants, which just utilize the first two parts of Eq.11 and Eq.23. We call them S-GDM-exp and S-GDM-mix respectively. The main purpose of doing this is to validate the feasibility of the added nuclear norm.

It should be emphasized that the results of MMD, AHISD, and SPDML on the two datasets have already been reported by the original authors and we use them directly. We have our own implementations of PML, LEML, S-GDM-exp, S-GDM-mix, GDM-exp, and GDM-mix respectively. In order to ensure a fair comparison, all the experimental settings in this paper are equivalent to the previous works [1, 14, 22, 23, 24, 25]. Tab. 1 shows the average classification accuracies of different methods on the ETH-80 and YTC datasets. The recognition scores of different algorithms on the Cambridge hand gesture dataset are shown in Tab. 2

TABLE I. AVERAGE CLASSIFICATION ACCURACIES OF DIFFERENT METHODS ON THE TWO DATASETS OBTAINED BY TEN-FOLD EXPERIMENTS

| **Methods** | **ETH-80** | **YTC** |
|---|---|---|
| MMD [35] | $85.72\pm 8.29$ | $62.90\pm 3.24$ |
| AHISD [36] | $77.25\pm 7.50$ | $63.70\pm 2.89$ |
| SPDML-Stein [28] | $90.50\pm 3.87$ | $61.57\pm 3.43$ |
| PML [1] | $90.00\pm 4.70$ | $66.83\pm 7.01$ |
| LEML [27] | $92.25\pm 2.19$ | $69.04\pm 3.84$ |
| **S-GDM-mix** | **92.00 ± 4.22** | **72.41 ± 3.20** |
| **GDM-mix** | **93.00 ± 3.50** | **73.26 ± 3.51** |
| **S-GDM-exp** | **92.50 ± 3.12** | **72.41 ± 3.26** |
| **GDM-exp** | **93.25 ± 3.34** | **73.69 ± 2.95** |

TABLE II. AVERAGE CLASSIFICATION ACCURACIES OF DIFFERENT METHODS ON CAMBRIDGE HAND GESTURE DATASETS OBTAINED BY TEN-FOLD EXPERIMENTS

| **Methods** | **CHG** |
|---|---|
| PML [1] | $85.67\pm 5.09$ |
| LEML [27] | $83.33\pm 2.82$ |
| **S-GDM-mix** | **92.44 ± 1.95** |
| **GDM-mix** | **92.89 ± 1.41** |
| **S-GDM-exp** | **92.89 ± 1.97** |
| **SMPM-exp** | **93.11 ± 1.80** |

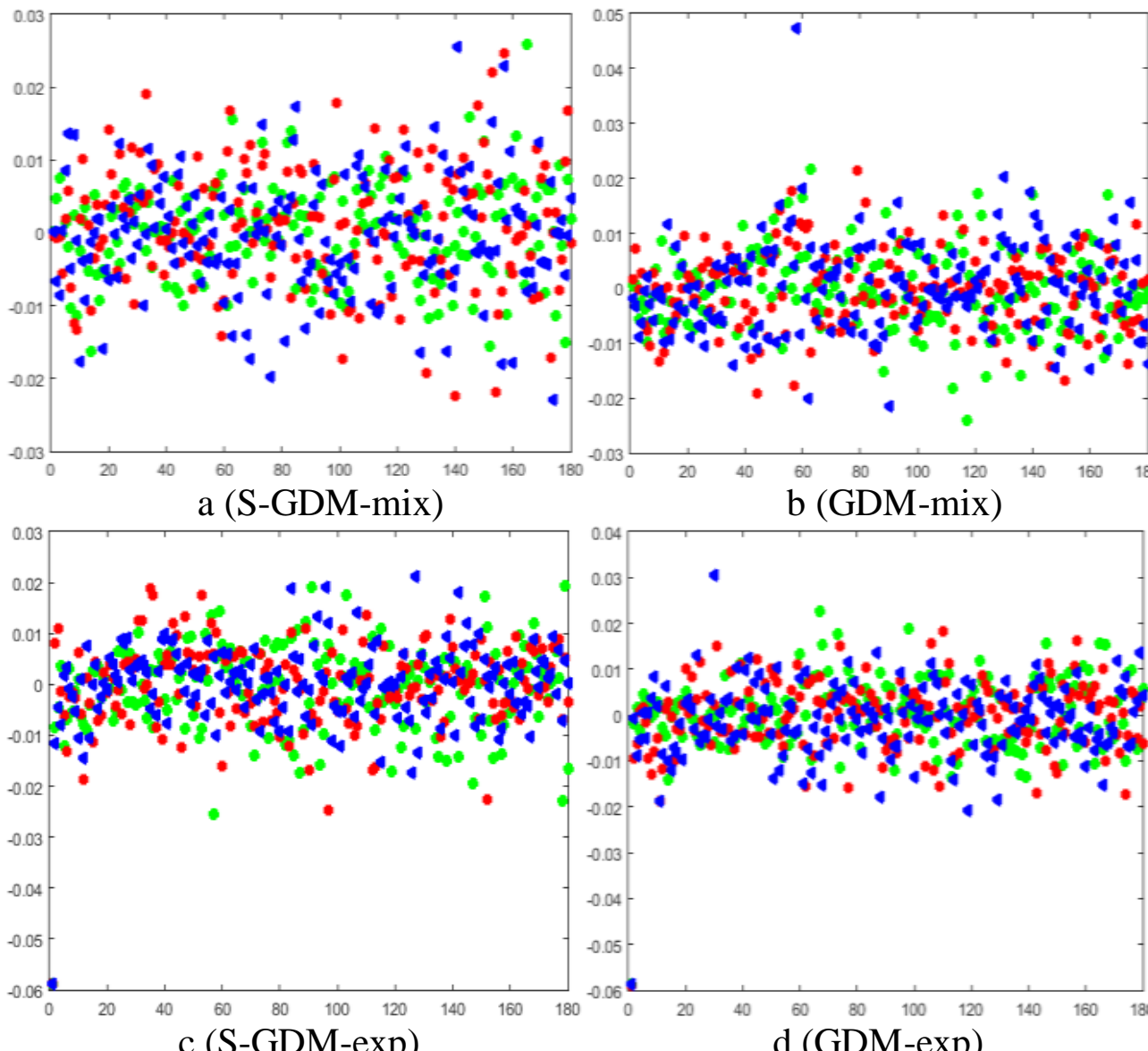

Fig. 3. Resulting data distributions

## C. *Results and Discussion*

Tab.1 and Tab.2 list the classification accuracies of different comparative methods on the different datasets. It is interesting to find the proposed method GDM-exp achieves better performance both in terms of recognition scores and standard derivations. As to its variant GDM-mix, it also shows a good classification performance on the three datasets, which demonstrates the effectiveness of the proposed methods. We can intuitively see S-GDM-mix and S-GDM-exp report the comparable results in the different classification tasks, and the performance of S-GDM-mix is inferior to S-GDM-exp, which is also noted for the proposed method and its simpler variant. This observation may denote the exponential operator imposes a more favorable constraint on the trade-off between the inter-class dispersion and intra-class compactness than the logarithm operator. Interestingly, the performance of the GDM-exp and GDM-mix are better than their simpler versions, which proves the added nuclear norm is feasible and effective.

To evaluate the merit of the nuclear norm, we make experiments on the YTC dataset to analyze the impact of these proposed algorithms on the data distribution of the resulting new linear subspaces. The experimental results are shown in Fig.3a, Fig.3b, Fig.3c, and Fig.3d respectively. From Fig.3 we can see that after utilizing these proposed algorithms, the discriminatory potential of the extracted features is improved, and we can intuitively find the dispersion of different classes is increased, the compactness of the intra-class is also increased when compared to the PML method. Nevertheless, when comparing the left column figures of Fig.3 with the right column figures, we can see these left column figures show a certain degree of compactness of the between-class. However, when we add the low rank constraint to the target matrix $P$, the distance between any pair of samples from different classes is increased and the distance between any pair of samples from the same class is decreased, which is of interest to identify that the added nuclear norm is working.

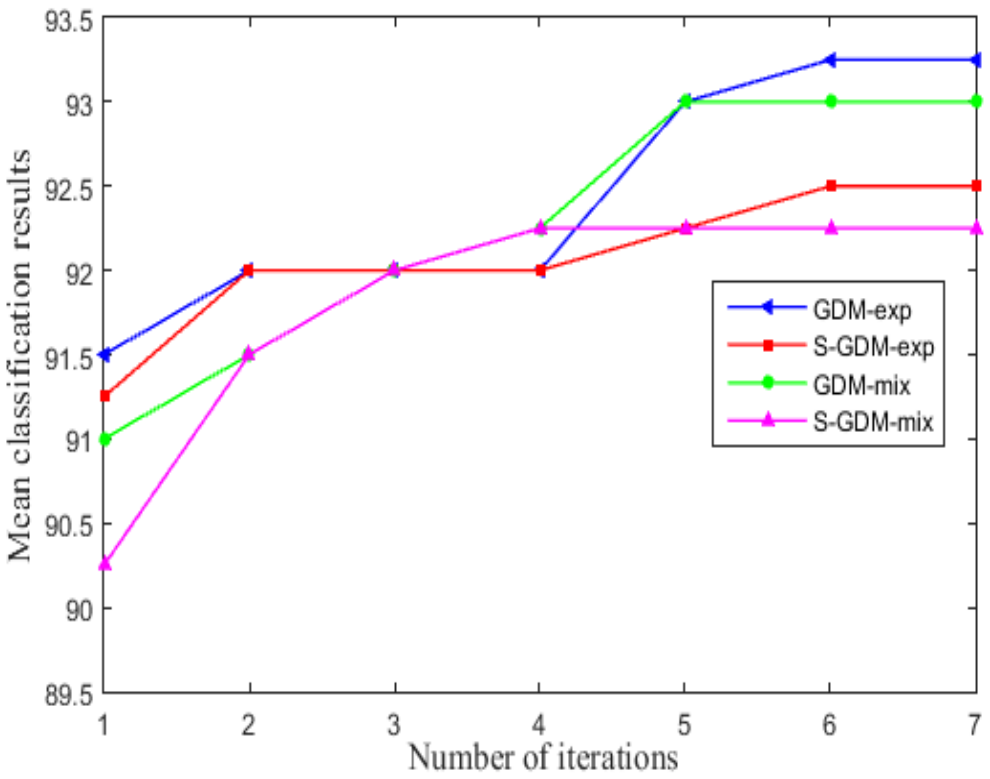

Fig. 4. The impact of the number of iterations to the classification rates of the proposed methods on ETH-80 dataset

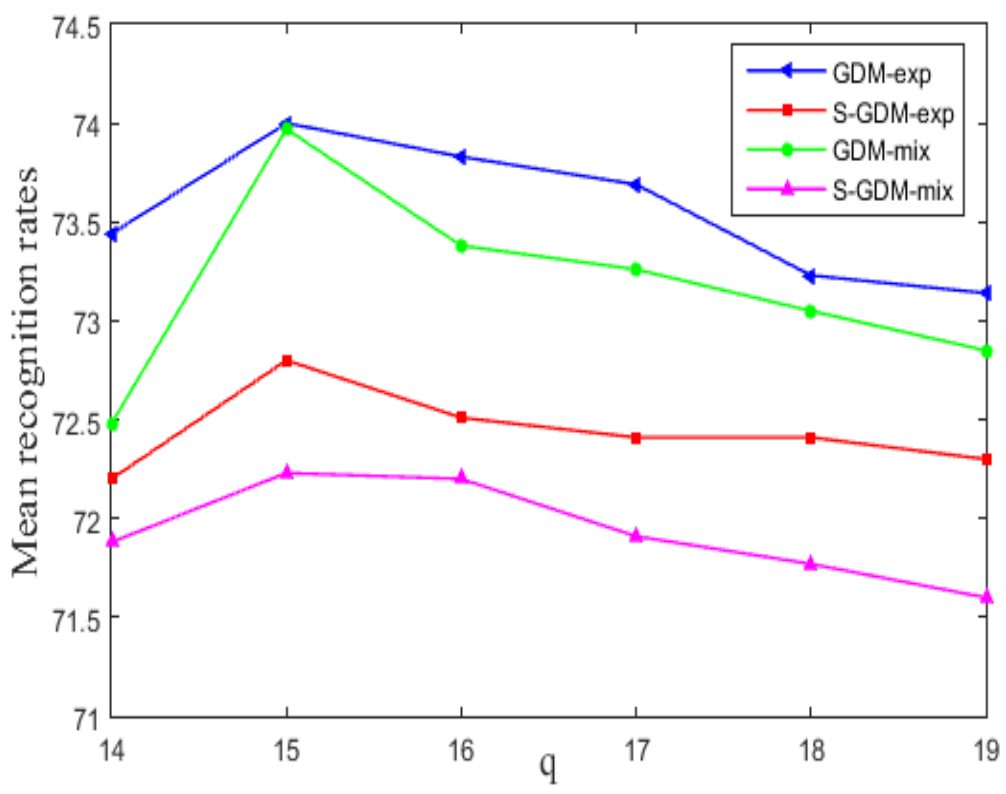

Fig. 5. The impact of the dimensionality of the original linear subspace to the classification rates of the proposed methods on YTC dataset

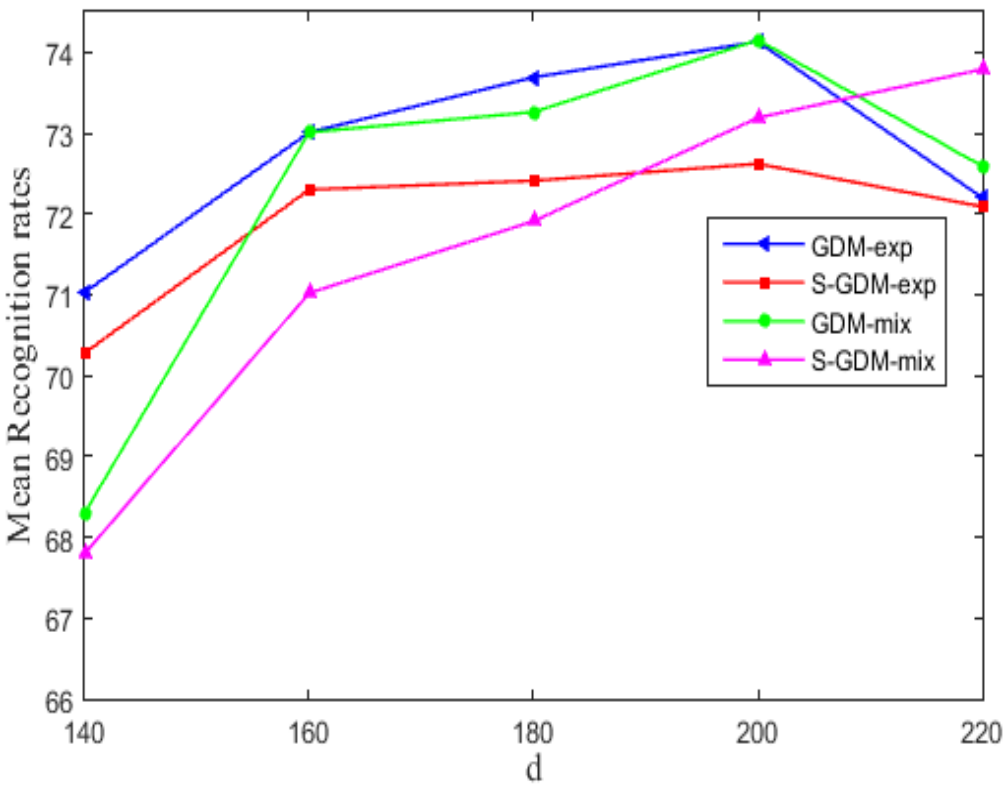

Fig. 6. The impact of the dimensionality of the target Grassmann manifold to the recognition results of the proposed methods on YTC dataset

It is difficult for us to provide a theoretical proof of convergence of the proposed method. Nevertheless, after a few iterations the proposed algorithm converges to a stable classification result (see Fig.4) on the ETH-80 dataset. This observation indicate the proposed method has a good convergence behavior. Meanwhile, this figure also help us to further prove that the added nuclear norm is useful and the proposed methods are efficient.

As is well known, the original Grassmann manifold is spanned by orthonormal linear subspaces, so it is important to choose a suitable $q$ . For this purpose, we make experiments on the YTC dataset. From Fig.5 we find the performance of the proposed algorithms tends to change smoothly with different $q$ , which confirms their robustness. For the YTC dataset, we found the most suitable value of $q$ is 17. Due to the space limitation, the detailed procedure of determining $q$ on the other two datasets is not elaborated. but for ETH-80 dataset and CHG dataset, we set $q=11$ and $q=17$ respectively.

Finally, we also evaluate the effect of different $d$ to the performance of the proposed methods on the YTC dataset. Fig.6 shows the average recognition results, and we can find with the increase of dimensionality, the recognition rates tend to change gently. However, an increase in dimensionality is often accompanied with an increase in computational cost and feature redundancy. In order to strike a balance between classification accuracy and computational cost, we set $d=180$ on the three datasets.

## V. CONCLUSION

In this paper, we proposed a new method for jointly performing dimensionality reduction and metric learning on Grassmann manifold, referred as Grassmannian Discriminant Maps (GDM). We devised a novel discriminant function for GDM to improve its ability to extract discriminatory information from the training set. We also introduce a simple extension of the GDM method, namely GDM-mix. We test it together with GDM on three different classification tasks. The extensive experimental results on three different datasets demonstrate the effectiveness of the proposed GDM and its simpler variant. The performance of GDM method outperforms SMDM-mix which shows the benefit of the exponential operator in the objective function. For future work, we plan to extend the proposed method to other Riemannian manifolds and to design effective discriminant functions for image set classification tasks defined on other datasets.